\documentclass[runningheads]{llncs}

\usepackage{accv}

\usepackage{accvabbrv}

\usepackage{graphicx}
\usepackage{booktabs}
\usepackage{wrapfig}
\usepackage{amsmath}
\usepackage{amssymb}
\usepackage{bm}
\usepackage{xcolor}
\usepackage{colortbl}
\usepackage{multirow}
\usepackage[accsupp]{axessibility}  

\usepackage{pifont}
\usepackage{adjustbox}

\definecolor{hl}{RGB}{245,246,242}

\def\blue#1{\textcolor{blue}{#1}}

\def\blue#1{\textcolor{blue}{#1}}

\usepackage{hyperref}

\usepackage{orcidlink}

\begin{document}

\title{DPL: Cross-quality DeepFake Detection via \\ Dual Progressive Learning} 

\titlerunning{DPL}

\author{Dongliang Zhang, Yunfei Li, Jiaran Zhou, Yuezun Li\thanks{Corresponding author (\url{liyuezun@ouc.edu.cn}).}}

\authorrunning{Yuezun Li et al.}

\institute{School of Computer Science and Technology, \\ Ocean University of China, Qingdao, China} 

\maketitle

\begin{abstract}
  Real-world DeepFake videos often undergo various compression operations, resulting in a range of video qualities. These varying qualities diversify the pattern of forgery traces, significantly increasing the difficulty of DeepFake detection. To address this challenge, we introduce a new Dual Progressive Learning (DPL) framework for cross-quality DeepFake detection. We liken this task to progressively drilling for underground water, where low-quality videos require more effort than high-quality ones. To achieve this, we develop two sequential-based branches to ``drill waters'' with different efforts. The first branch progressively excavates the forgery traces according to the levels of video quality, \ie, time steps, determined by a dedicated CLIP-based indicator. In this branch, a Feature Selection Module is designed to adaptively assign appropriate features to the corresponding time steps. Considering that different techniques may introduce varying forgery traces within the same video quality, we design a second branch targeting forgery identifiability as complementary. This branch operates similarly and shares the feature selection module with the first branch. Our design takes advantage of the sequential model where computational units share weights across different time steps and can memorize previous progress, elegantly achieving progressive learning while maintaining reasonable memory costs. Extensive experiments demonstrate the superiority of our method for cross-quality DeepFake detection.
  
\end{abstract}

\section{Introduction}
\label{sec:intro}

The rapid advancement of deep learning technologies has significantly propelled the evolution of \emph{Deepfakes}, a generative technique capable of creating highly realistic faces within videos. Misusing DeepFake technique poses serious societal concerns, including forging fake news, economic fraud, and making illusions of public figures~\cite{DBLP:journals/snam/AimeurAB23,chesney2019deepfakes,deepfakes_trick_or_threat}. 
In response to this problem, numerous forensics methods have emerged to detect DeepFakes~\cite{qian2020thinking,Lee_An_Woo_2022,capsule,recce_cvpr2022}. These methods, typically developed on deep neural networks (DNNs), have shown promising and even near-perfect performance on standarded datasets \cite{rössler2019faceforensics,Celeb_DF_cvpr20,FFIW10k,dfdc}, seemingly indicating that the DeepFake problem has been successfully addressed. 

However, real-world DeepFake videos often undergo various compression operations, resulting in varying video quality. Social platforms (\eg, TikTok, Instagram) and video platforms (\eg, YouTube, YouKu) usually compress the uploaded videos to reduce broadcasting overhead. These compression operations disturb the pattern of forgery traces, significantly degrading the performance of existing methods. Thus, exploring the practical feasibility of DeepFake detection against quality variation is pressing.

To address this concern, several efforts have been proposed~\cite{Lee_An_Woo_2022,QAD_iccv2023,le2022add}. Noticing that existing attempts are usually validated on high-quality DeepFake videos, some methods~\cite{Lee_An_Woo_2022,le2022add} have shifted their focus to low-quality DeepFake videos. However, these methods do not investigate the generalization detection across various qualities. More recently, QAD~\cite{QAD_iccv2023} has extended to achieve cross-quality DeepFake detection, by employing specific learning strategies (\eg, contrastive learning and collaborative learning) to capture generic forgery traces across different qualities. Nonetheless, when the quality discrepancy is substantial, these models may reach their capacity limits and struggle to capture the generic forgery traces. While ensumbling multiple models may overcome these limits, it introduces significant computational overhead, hindering their practical applications.

\begin{wrapfigure}[15]{r}{0.5\linewidth}
\vspace{-0.6cm}
    \centering
    \includegraphics[width=\linewidth]{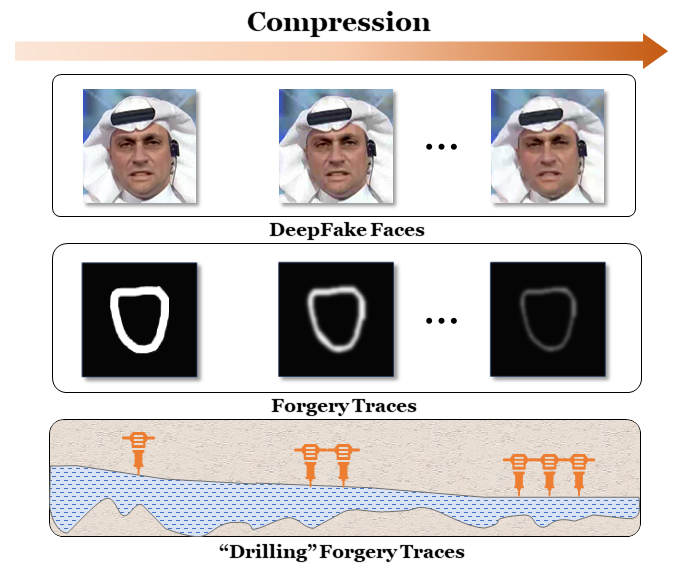}
    \vspace{-0.6cm}
    \caption{Different image quality}
    \label{fig:PL1}
\end{wrapfigure}


In this paper, we rethink this problem and propose a Dual Progressive Learning (DPL) framework for cross-quality DeepFake detection. Our method is designed based on a novel perspective: we liken forgery traces to underground water and the process of DeepFake detection to drilling for water. In high-quality videos, forgery traces are likely ``near the surface'' and thus easily ``drilled''. However, in low-quality videos, these traces are greatly disturbed by compression, \ie, ``concealed deeper'', requiring more effort to reveal them (see \cref{fig:PL1}). 
Thus, a natural question arises ``\textbf{Can we design a dynamic model that adaptively assigns efforts for different DeepFake videos?}''

To achieve this, we first develop a Visual Quality Guided Progressive Learning (VQPL) branch. This branch is designed based on a sequential model, which first estimates the video quality levels as different time steps and progressively engages computational unit (\eg, GRU~\cite{gru}) to excavate forgery traces, with lower quality requiring more time steps and vice versa. 
Specifically, the video quality level is estimated by a new CLIP-based Visual Quality Indicator (VQI). Considering that each time step in progressive learning likely focuses on different views, \ie, ``the depth of drilling'', we propose a Feature Selection Module (FSM), which assigns appropriate features for computational units at different time steps.
The benefit of this design is that the sequential model is inherently dynamic with weight-sharing computational units and can memorize the previous progress, reducing the computational cost while increasing the model diversity, similar to ensumble multiple models.  

\begin{wrapfigure}[15]{r}{0.5\linewidth}
\vspace{-0.6cm}
    \centering
    \includegraphics[width=\linewidth]{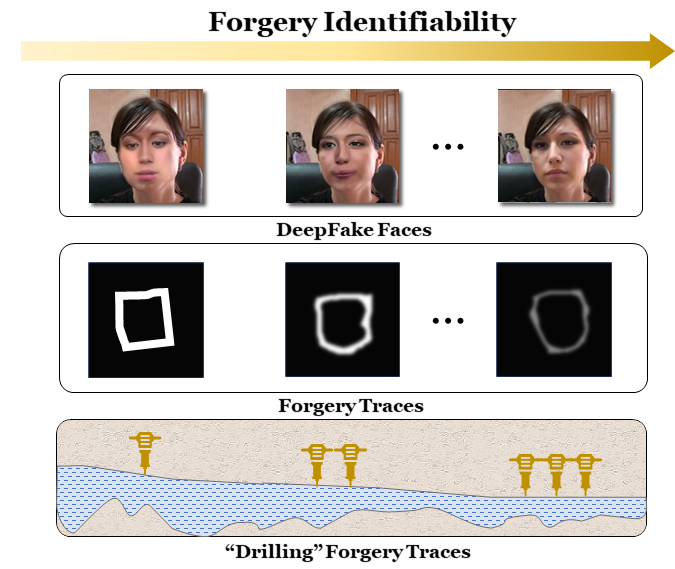}
    \vspace{-0.6cm}
    \caption{Different forgery quality}
    \label{fig:PL2}
\end{wrapfigure}

Besides varying video quality, the level of forgery identifiability is another implicit obstacle hindering the detection performance. As shown in \cref{fig:PL2}, while these DeepFake faces have the same compression level, their forgery identifiability differs. This challenges the proposed VQPL branch. To compensate, we introduce another Forgery Identifiability Guided Progressive Learning (FIPL) branch, which borrows the same spirit of VQPL to handle varying levels of forgery identifiability. This branch shares the same feature selection module and employs computational units controlled by a new CLIP-based Forgery Identifiability Indicator (FII). We then fuse the features of these two branches for DeepFake identification.

To effectively train the proposed DPL framework, we introduce a two-stage pipeline with a Proximal Policy Optimizer (PPO) algorithm \cite{ppo}, that is applied on dedicated designed objective functions.
Extensive experiments are conducted on many public datasets and compared to several state-of-the-art methods. The results show that our method outperforms others, demonstrating the efficacy of our method in cross-quality DeepFake detection.

Our contributions can be summarized as follows: 
\begin{itemize}
    \item We propose a new Dual Progressive Learning framework (DPL) for cross-quality DeepFake detection. In contrast to existing methods, we formulate this task as progressively excavating forgery traces according to varying levels of video quality and forgery identifiability.

    \item For both branches, we introduce a shared Feature Selection Module (FSM) that can flexibly assign features to computational units. For each branch, we design CLIP-based indicators to determine the time steps.

    \item We describe a two-stage training pipeline with devoted designed objective functions, to effectively train the proposed DPL framework.

    \item Extensive experiments are conducted on several public datasets, comparing our method to many state-of-the-art methods. We also thoroughly study the effect of each component, revealing interesting insights for future research. 
\end{itemize}

\section{Related Work}
Deepfakes have become a highly critical issue due to the significant threats they pose to social security and privacy. To counteract the risk, a large number of deepfake detection methods have been proposed~\cite{DBLP:journals/corr/abs-2207-08380,qian2020thinking,capsule,Chen_2022_ACCV,Lee_An_Woo_2022,Liu_2022_ACCV, multi_scale_accv2022}. These methods are typically developed on deep neural networks (DNNs) and attempt to uncover the forgery traces concealed in various aspects, such as physiological signals \cite{DBLP:journals/corr/abs-2207-08380,korshunov2018deepfakesnewthreatface,DeepRhythm}, frequency domain \cite{qian2020thinking,srm,spsl}, blending boundaries \cite{face-xray,sbi,SLADD}, and high dimensional feature spaces derived by networks \cite{capsule,Agarwal_2024_CVPR,LSDA_cvpr2024}. Although these approaches have demonstrated their promising capacity on public datasets, they still face great challenges when applying them in real-world scenarios.

Unlike standard datasets, the real-world DeepFake videos are more diversified, greatly degrading the performance of existing methods. Several recent methods have been proposed to enhance the generalization ability of detection, mainly concentrating on the scenarios of different manipulations~\cite{ucf_iccv2023,LSDA_cvpr2024,sbi} and different datasets~\cite{face-xray,altFreezing,Liu_2022_ACCV}. Besides these scenarios, the variation in video quality is also a crucial obstacle in DeepFake detection. For instance, social / video platforms usually apply compression operations to reduce memory costs, resulting in varying forgery traces and degrading detection performance. The works of \cite{le2022add,Lee_An_Woo_2022,QAD_iccv2023} focus on this problem and attempt to expose DeepFake videos with various qualities. In this paper, we describe a new method based on a novel perspective: we treat this task as ``drilling'' water and develop a Dual Progressive Learning (DPL) framework to enhance the detection ability under cross-quality scenarios.


\section{Method}
\begin{figure}[tb]
    \centering
    \includegraphics[width=1\linewidth]{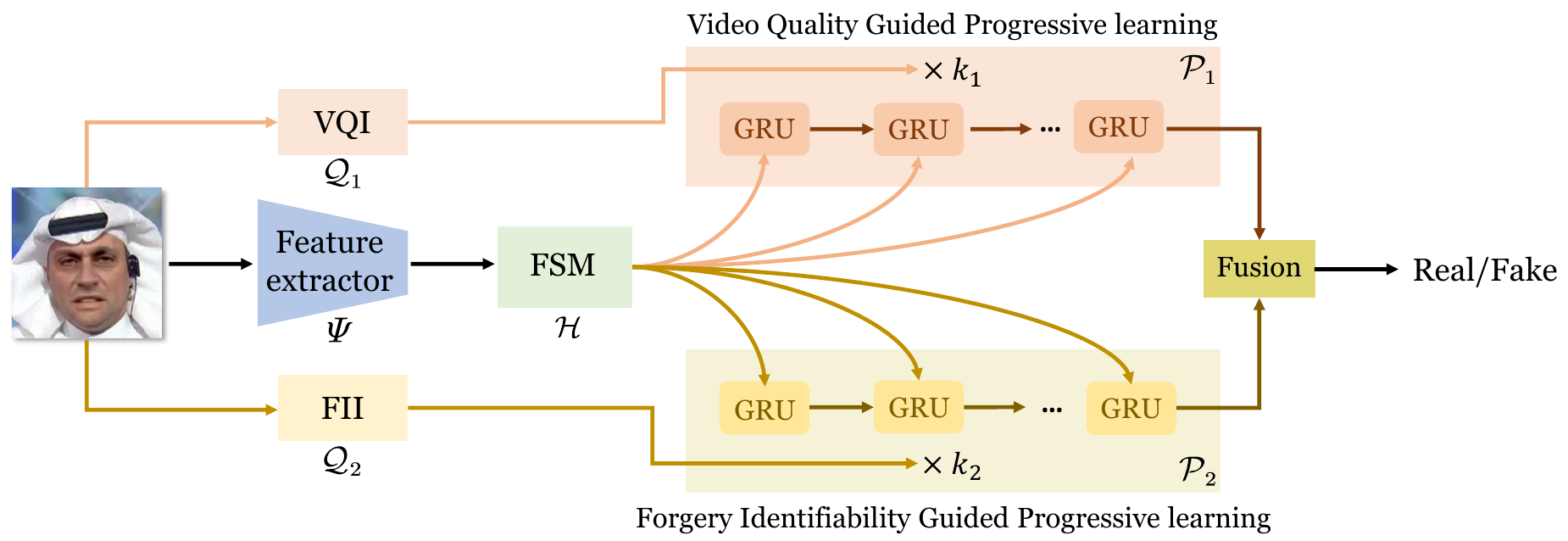}
    \caption{Overview of proposed DPL framework.}
    \label{fig:framework_fig}
    \vspace{-0.5cm}
\end{figure}

This paper describes a new Dual Progressive Learning (DPL) framework to detect DeepFakes across different video qualities. Specifically, we develop a Visual Quality Guided Progressive Learning (VQPL) branch to extract forgery traces progressively with the instruction of a Visual Quality Indicator (VQI) (\cref{sec:vqpl}). Moreover, we describe a Forgery Identifiability Guided Progressive Learning (FIPL) branch to overcome the difficulties introduced by forgery identifiability (\cref{sec:fipl}). These two branches collaborate to capture the generic forgery traces by fusing the output features of each branch. Then we introduce specific objectives and training procedures devoted to the proposed framework (\cref{sec:training}). The overall framework of our method is illustrated in \cref{fig:framework_fig}.


\subsection{Visual Quality Guided Progressive Learning}
\label{sec:vqpl}
This branch consists of four key components: a feature extractor, a feature selection module, a sequential-based detection module, and a video quality indicator. 
Given an input face image $\mathbf{x} \in [0,255]^{H \times W \times 3}$, we first extract the feature $f \in \mathbb{R}^{h \times w \times c}$ using a feature extractor $\Psi$ (\ie, backbone network). This feature serves as the input for the sequential-based detection module, a recurrent network $\mathcal{P}_1$ with GRUs~\cite{gru}. The maximum time step of $\mathcal{P}_1$ is represented by the visual quality level of input face image $\mathbf{x}$, estimated by a video quality indicator $\mathcal{Q}_1$. Denote $k_1 = \mathcal{Q}_1(\mathbf{x}) \in \{ 1, ..., K_1\}$ as the estimated quality level. Then the detection module $\mathcal{P}_1$ is executed $k$ times with the following formulas
$h_t = \mathcal{P}_1(f_t, h_{t-1})$, where $f_t$ denotes the input feature at time step $t \in \{1,... k_1 \}$, $h_{t-1}$ and $h_{t}$ are the hidden state and current output (a hidden state for next time step). Note that $h_{0}$ is the initial state and $h_{k_1}$ is the final output of this detection module. Then we fuse the outputs of all time steps as the final feature of this branch $f_{\textrm{VQ}} = \sum_t h_t$.

Intuitively, the feature $f$ can be employed as the input for all time steps, \ie, $f_t = f, t \in \{1,... k_1 \}$. However, considering the focus on progressive learning likely varies. For example, the early time steps are likely capable of capturing superficial traces while the late time steps should be responsible for more challenging traces. 
Therefore, we design a Feature Selection Module $\mathcal{H}$, which can process the feature $f$ for different time steps, \ie, $f_t = \mathcal{H}_t (f), t \in \{1,... k_1 \}$.
\begin{wrapfigure}[17]{r}{0.5\textwidth}
    \centering
    \includegraphics[width=\linewidth]{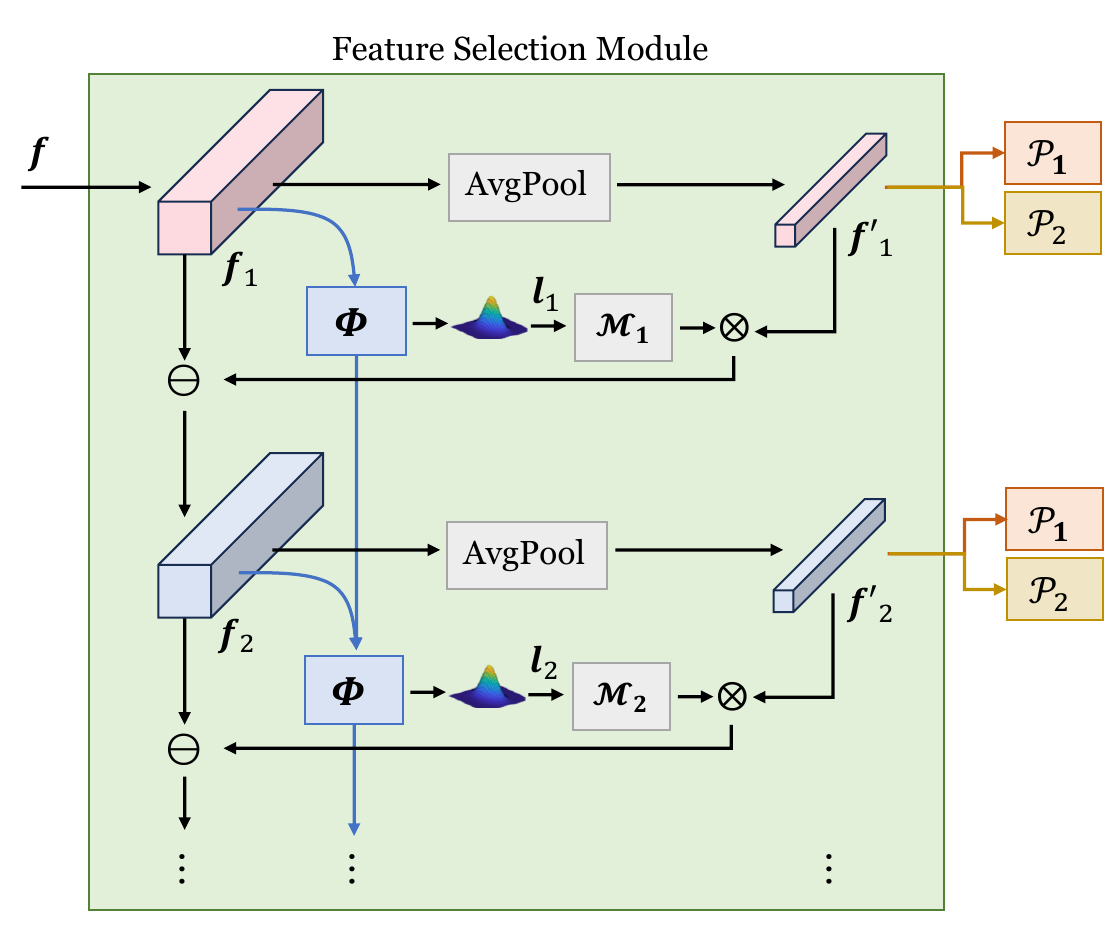}
    \vspace{-0.6cm}
    \caption{Overview of the feature selection module}
    \label{fig:fsm}
\end{wrapfigure}



\smallskip
\noindent\textbf{Feature Selection Module.}
The core of this module is another recurrent network $\Phi$, which is employed to recursively process the input feature of $\mathcal{P}_1$ for the current time step.
Specifically, given the input feature $f_{t-1}$ at time step $t-1$, we derive a mask based on this feature and perform a masking operation to refine $f_{t-1}$ as $f_t$.
By learning masking operations recursively, the features corresponding to challenging forgery traces will be retained for the later time steps, leading to progressive learning.

As shown in \cref{fig:fsm}, the network $\Phi$ takes as input the features $f_{t} \in \mathbb{R}^{h \times w \times c}$ and generate a mean $\mu_t$ and a standard variation $\sigma_t$ to form a Gaussian distribution $\mathcal{N}(\mu_t, \sigma_t)$. This distribution outlines the probability of masking positions. This process can be defined as $\mu_t, \sigma_t, g_{t} = \Phi(f_{t}, g_{t-1})$, where $g_{t-1}$ and $g_{t}$ are the hidden states for the previous and current time step. Then we create a mask $\mathcal{M}_t \in \mathbb{R}^{h \times w \times c}$ with the same size as $f_t$. Given the distribution $\mathcal{N}(\mu_t, \sigma_t)$, we can sample a starting position $l_t$ and scale it to the range $[0, c]$. Then we let $\mathcal{M}^{(:,:,i)}_t = 0$ if $i \in [l_t, l_t+\delta_t]$, otherwise $\mathcal{M}^{(:,:,i)}_t = 1$. Note that $\delta_t$ is a hyper-parameter that represents the size of the retained region. 
Then we obtain $f'_t$ by performing average pooling on $f_t$ and perform element-wise multiplication with $\mathcal{M}_t$. The results are subtracted from $f_t$ to retain suitable features for the next time step. The process of feature selection can be defined as $f_{t+1} = f_t - \mathcal{M}_t \otimes f'_t$. Then we perform average pooling on $f_{t+1}$ to obtain $f'_{t+1}$, the input feature for the time step $t+1$ in $\mathcal{P}_1$. Note that the feature at the first time step is $f_1 = f$.

\begin{wrapfigure}[14]{r}{0.4\textwidth}
\vspace{-0.5cm}
    \centering
    \includegraphics[width=1\linewidth]{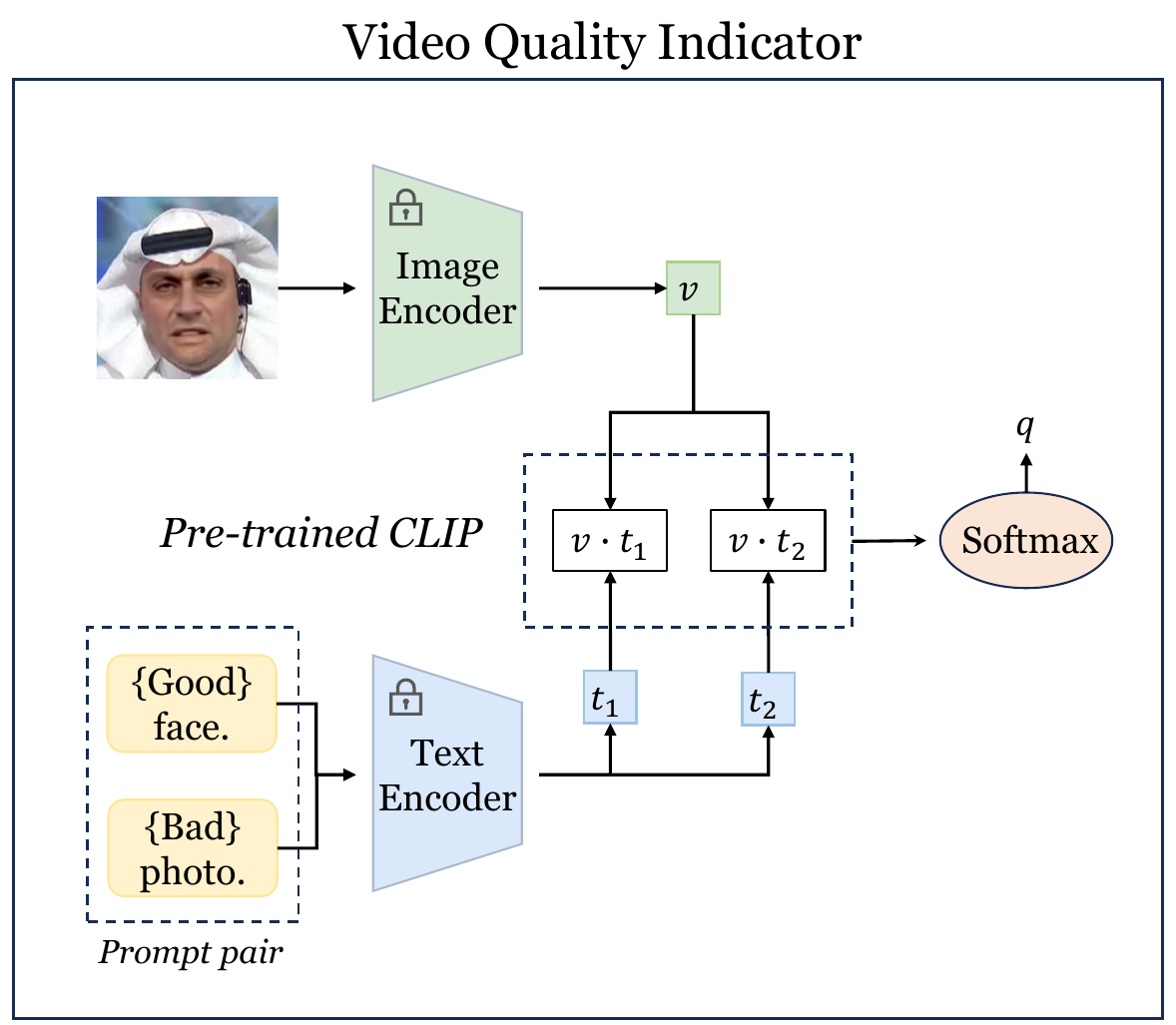}
    \caption{Overview of VQI.}
    \label{fig:VQI_architecture}
\end{wrapfigure}
\smallskip
\noindent\textbf{Video Quality Indicator.}
\label{sec:vqi}
Benefiting from the powerful semantic prior knowledge of vision-language models (VLMs)~\cite{clip_icml2021, vlm_introduction}, we develop a CLIP-based video quality indicator $\mathcal{Q}_1$ to estimate the visual quality levels of the input face image, severing as the time steps in sequential-based detection module $\mathcal{P}_1$. Inspired by~\cite{clip_iqa_iccv2023}, we adopt the paired text prompt of \texttt{[``Good photo.'', ``Bad photo.'']} to estimate the image quality. Specifically, given the input face image $\mathbf{x}$, we obtain the visual embedding from the image encoder as $v = \mathcal{Q}^{IE}_1(\mathbf{x})$. Denote the text embedding from the text encoder as $t_1 = \mathcal{Q}^{TE}_1(\texttt{[``Good photo.'']})$ and $t_2 = \mathcal{Q}^{TE}_1(\texttt{[``Bad photo.'']})$. Then we can obtain the quality score $q$ as
\begin{equation}
\small
    \begin{aligned}
         & s_i=\frac{v\odot t_i}{\left\| x \right\| \cdot \left\| t_i \right\|}, i\in \left\{ 1,2 \right\} \\
         & q=\frac{\exp(s_1)}{\exp(s_1+s_2)}, \\
    \end{aligned}
    \label{eq:vqi}
\end{equation}
where $s_i$ is the cosine similarity between visual and corresponding text embedding, which is performed \texttt{softmax} as the final score. A larger $q$ indicates the image $\mathbf{x}$ has better quality and vice versa.

Generally, low-quality videos undergo compression operations, disrupting the pattern of forgery traces. Thus they exhibit unnoticeable forgery artifacts, making them more difficult to detect than high-quality videos. To this end, we should assign more steps for low-quality videos. Since the quality scores are continuous, we should discretize them to represent time steps. Specifically, we first estimate the quality scores for all images in the training set and divide the scores into $K_1$ bins, ensuring that the range of each bin can include a similar number of images. Inspired by Likert-scale~\cite{DBLP:journals/tip/GhadiyaramB16,liqe_cvpr2023}, we establish $K_1=5$ quality levels using $k\in \left\{ 1,2,3,4,5 \right\} =\left\{ \mathrm{``perfect"},\mathrm{``good"},\mathrm{``fair"},\mathrm{``poor"},\mathrm{``bad"} \right\} $. It is worth noting that a smaller bin index represents better image quality, corresponding to assigning fewer time steps. The overview of VQI is shown in \cref{fig:VQI_architecture}.




\subsection{Forgery Identifiability Guided Progressive Learning}
\label{sec:fipl}
This branch is complementary to the VQPL branch, handling the variation in the forgery identifiability. Specifically, it shares the same feature extractor $\Psi$ and feature selection module $\mathcal{H}$. But it employs another sequential-based detection module $\mathcal{P}_2$ to progressively excavate forgery traces based on forgery identifiability levels, estimated by a new CLIP-based Forgery Identifiability Indicator (FII). 
Denote $k_2 \in \{ 1, ..., K_2 \}$ as the identifiability level of image $\mathbf{x}$. Similar to $\mathcal{P}_1$, the module $\mathcal{P}_2$ is executed $k_2$ times, described by $w_t = \mathcal{P}_2(f_t, w_{t-1})$, where $f_t$ is the input feature processed by feature selection module $\mathcal{H}$ and $w_{t}$ is the hidden state. Note that $w_k$ is the final output of this branch. Then we fuse the outputs of all time steps as $f_{\textrm{FI}} = \sum_t w_t$.

\begin{wrapfigure}[14]{r}{0.4\textwidth}
    \vspace{-0.5cm}
    \centering
    \includegraphics[width=1\linewidth]{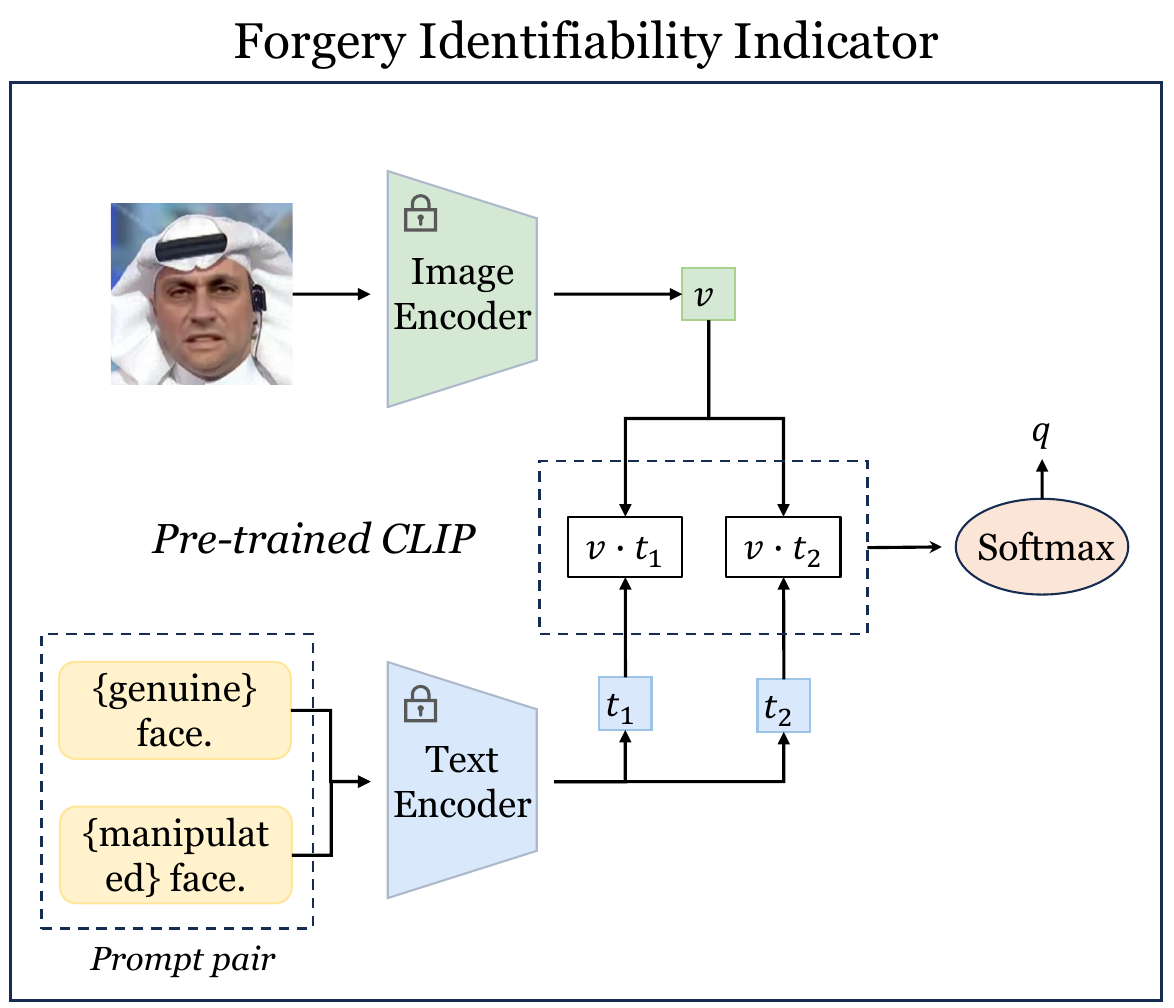}
    \caption{Overview of FII.}
    \label{fig:FII_architecture}
\end{wrapfigure}

\smallskip
\noindent\textbf{Forgery Identifiability Indicator.}
This indicator, denoted as $\mathcal{Q}_2$, shares the same CLIP model and estimation pipeline used in VQI. However, we design a different paired text prompt \texttt{[``manipulated face.'', ``genuine face.'']}, which can represent whether the given image $\mathbf{x}$ looks like being manipulated, reflecting the forgery identifiability levels. Similar to VQI, we also employ the \cref{eq:vqi} to calculate the score and use the same strategy to discretize the scores into $K_2 = 5$ levels. A smaller $k_2$ denotes the forgery is more identifiable, thus requiring less effort to detect, otherwise requiring more effort. The overview of VQI is shown in \cref{fig:FII_architecture}.

\subsection{Objective and Training}
\label{sec:training}
At the beginning of training, the framework likely struggle to identify forgery traces, making the recurrent network $\Phi$ in feature selection module difficult to converge. Thus, we describe a two-stage training pipeline that trains feature selection module and dual branches respectively. In the first stage, we enhance the feature extraction capacity by training the dual branches without $\Phi$. Once the whole network has developed adequate feature extraction capacity, we shift our focus to training $\Phi$ within the feature selection module, keeping the rest of the network fixed during the second stage.

\smallskip
\noindent{\bf Stage I.}
In this stage, we replace $\Phi$ with a randomizer, where the position $l_t$ for creating the mask is randomly sampled from a uniform distribution, indicating that the input feature for each time step is randomly masked without using $\Phi$. 

In training, we employ the cross-entropy loss $\mathcal{L}_{\textrm{CE}}$, enhanced by the Focal Loss~\cite{focal_loss} to learn the discriminative features between real and fake faces. 
Moreover, we introduce a regularization loss $\mathcal{L}_{\textrm{REG}}$ to enhance the learning of hard samples, the samples are often misclassified. At early time steps, the network may easily misclassify these samples due to insufficient learning capacity at the beginning of progressive learning. Therefore, we can make the network refrain from making decisions (\ie, real or fake) in the early time steps, deferring these decisions to the later time steps. Specifically, we use the entropy value to assess whether a decision is made. A larger entropy value denotes the confidence for real and fake classes is similar, indicating the authenticity is not determined. This loss can be defined as
\begin{equation}
    \begin{aligned}
        \mathcal{L}_{\textrm{REG}} = - \sum_{\mathbf{x}_i} \Pi(\mathbf{x}_i) \; \left [ \sum_{j \in \{0,1\}} c^j_{i} \log (c^j_{i}) \right ],
    \end{aligned}
\end{equation}
where $\Pi(\mathbf{x}_i)$ is an indicator function, which is equal to $1$ if $\mathbf{x}_i$ is a hard sample, $0$ otherwise, $c^j_i$ denotes the confidence of $j$-th class on $\mathbf{x}_i$, where $j \in \{0,1\}$ represents real and fake respectively. Thus, in this stage, the overall objective function can be defined as
\begin{equation}
\begin{aligned}
    \mathcal{L}_1 =  \mathcal{L}_{\textrm{CE}} + \mathcal{L}_{\textrm{REG}}.
\end{aligned}
\end{equation}

\smallskip
\noindent{\bf Stage II.}
In this stage, we revoke the network $\Phi$ but fix other parameters. 
To effectively train $\Phi$, we employ the Proximal Policy Optimizer algorithm (PPO)~\cite{ppo} to maximize the total reward. 
The motivation is that the network $\Phi$ should assist VQPL and FIPL to progressively improve the performance along with the time steps increasing. Denote $c_t$ as the predicted probability of input image $\mathbf{x}$ for the ground truth label and $r_t = c_t - c_{t-1}$ as the rewards. The goal of this stage is to maximize the total rewards as $\max_{\Phi} \sum_{t=2}{r_t}$.
Inspired by~\cite{GFNets_neuips2020}, we adopt three objective terms for this goal. Specifically, we employ a reward learning term $\mathcal{L}_{\textrm{REW}}$~\cite{ppo} which instructs the optimization direction of $\Phi$ as
\begin{equation}
\begin{aligned}
    \mathcal{L} _{\mathrm{REW}}=-\mathbb{E} _t\left[ \min \left( \frac{p_{\Phi}\left( l_t|f_t \right)}{p_{\Phi ^{\prime}}\left( l_t|f_t \right)},
    \mathrm{clip}\left( \frac{p_{\Phi}\left( l_t|f_t \right)}{p_{\Phi ^{\prime}}\left( l_t|f_t \right)},1-\epsilon ,1+\epsilon \right) \right) \mathcal{A} _t \right] ,
\end{aligned}
\end{equation}
where $p_{\Phi}\left( l_t|f_t \right)$ represents the probability of sampling $l_t$ from $\Phi$, while $p_{\Phi ^{\prime}}\left( l_t|f_t \right)$ denotes the probability of sampling $l_t$ from $\Phi ^{\prime}$. Note that $\Phi$ and $\Phi'$ indicate the state after and before the update. $\mathcal{A}_t$ is an indicator at time step \(t\), measuring the increase of reward compared to a standard value. $\mathcal{A}_t > 0$ indicates the current state of $l_t$ can increase the reward, thus increasing $p_{\Phi}\left( l_t|f_t \right)$ and vice versa.

Specifically, $\mathcal{A}_t = -\mathcal{V}(f_t)  + \mathcal{R}_{t}$, where $\mathcal{R} _t=\sum_t{r_t}$ denotes the sum of rewards at all steps and $\mathcal{V}(f_t)$ is the standard obtained using a sub-network $\mathcal{V}$ with a squared error loss $\mathcal{L}_{\textrm{SE}}$ as
\begin{equation}
\begin{aligned}
    \mathcal{L}_{\textrm{SE}} = \lVert \mathcal{V}(f_t) - \mathcal{R}_{t} \rVert^{2}.
\end{aligned}
\end{equation}
Minimizing this term can allow the sub-network $\mathcal{V}$ to learn the practical rewards. 
Moreover, in case the $l_t$ is only sampled in a small range, we describe a term $\mathcal{L}_{\textrm{EN}}$ to measure the entropy of the distribution, where a larger value denotes the distribution is flatter. The overall objective in this stage is defined as
\begin{equation}
\begin{aligned}
    \mathcal{L}_2 = \mathcal{L}_{\textrm{REW}} + \mathcal{L}_{\textrm{SE}} -  \mathcal{L}_{\textrm{EN}}.
\end{aligned}
\end{equation}

\section{Experiments}

\subsection{Experimental Settings}

\smallskip
\noindent\textbf{Datasets.}
Our method is validated on widely used deepfake datasets: FaceForensics++ (FF++)~\cite{rössler2019faceforensics}, CelebDFv2 (CDFv2)~\cite{Celeb_DF_cvpr20}, FaceShifter (FSH)~\cite{FaceShifter} and FFIW10K~\cite{FFIW10k}. FF++ datasets comprise 1000 original videos and 4000 fake videos produced in four manipulation methods, Deepfakes (DF), Face2Face (F2F), FaceSwap (FS), and NeuralTextures (NT), with two widely used quality versions: c23 (lightly compressed), and c40 (heavily compressed). CDFv2 \cite{Celeb_DF_cvpr20} is a recent high-quality dataset with more than 5600 DeepFake videos. FSH \cite{FaceShifter} contains 1000 original videos and 1000 fake videos. FFIW10k consists of 10000 fake videos with an average of three human faces in every frame.



\smallskip \noindent\textbf{Implementation Details.}
\label{sec:implementation_detail}
We employ ConvNeXt-T \cite{convnext} as our feature extractor. During the training phase, we employ random JPEG compression for data augmentation. The size of the retained region is set $\delta_t=c/3$. The DPL framework is trained using Adam optimizer \cite{adam} with a learning rate of 5e-5. The batch size is 32 and the total training epoch is 10, where 5 epochs are used in stage I and 5 epochs are used in stage II. The input image size is 224 $\times$ 224.

\subsection{Results}
Our method is compared to eight state-of-the-art methods, including F3Net (ECCV'20) \cite{qian2020thinking}, SRM (CVPR'21) \cite{srm}, SPSL (CVPR'21) \cite{spsl}, SIA (ECCV'22) \cite{sia_eccv2022}, RECCE (CVPR'22) \cite{recce_cvpr2022}, UIA-ViT (ECCV'22) \cite{uia-vit_eccv2022}, QAD (ICCV'23) \cite{QAD_iccv2023}, UCF (ICCV'23) \cite{ucf_iccv2023}. Note that all methods are trained on FF++(c23) and evaluated by the Area Under Curve (AUC) metric for a fair comparison.

\smallskip
\noindent\textbf{Cross-quality Evaluation.}
In this setting, the methods are evaluated on the test set for each manipulation method DF, F2F, FS, and NT, under random JPEG compression.
The results are presented in \cref{tab:within_domain_evaluation}, showing that our method achieves superior performance compared to other methods in most cases and performs the best on average, improving the AUC by approximately $1\%$ compared to the second-best method QAD.

Moreover, we further evaluate these methods in a more challenging setting, where the train set remains FF++(23) and the test set is FF++(40). The test set is subjected to the same random JPEG compression. The results, shown in \cref{tab:cross_quality_evaluation}, indicate that our method consistently outperforms the others, achieving the highest average score and improving the AUC by about $1\%$ over the second-best method UIA-ViT.

\begin{table}[!tb]
    \centering
    \caption{\textbf{Cross-quality Evaluation (AUC \%)}. Train set: FF++(\textbf{c23}). Test set: FF++(\textbf{c23}) with random JPEG compression. The \textbf{best} and \underline{second-best} scores are highlighted in bold and underlined.}
    \label{tab:within_domain_evaluation}
    \vspace{-0.3cm}
    \setlength{\tabcolsep}{0.3cm}
    \begin{tabular}{lccccc}
        \toprule
        Methods & DF & F2F & FS & NT & Avg. \\
        \midrule
        \multicolumn{6}{c}{\textbf{Random JPEG compression on c23 test set}} \\
        \midrule
        F3Net (ECCV'20) \cite{qian2020thinking} &  94.93& 91.75&  93.22&83.47&  90.84\\
        SRM (CVPR'21) \cite{srm} & 96.16& \underline{95.09} & 95.67& 84.93&92.96\\
        SPSL (CVPR'21) \cite{spsl} &  95.60& 95.00&  93.93& \underline{85.96} &  92.62\\
        SIA (ECCV'22) \cite{sia_eccv2022} & 93.86& 94.06& 93.73& 81.73&90.83\\
        RECCE (CVPR'22) \cite{recce_cvpr2022} &  95.31& 94.17&  95.18&84.14&  92.20\\
        UIA-ViT (ECCV'22) \cite{uia-vit_eccv2022} & 95.94& 91.40& 93.02& 80.33&90.17\\
        QAD (ICCV'23) \cite{QAD_iccv2023} &  \underline{97.56} & 94.71&  \textbf{97.11} &84.55& \underline{93.48} \\
        UCF (ICCV'23) \cite{ucf_iccv2023} &  96.87& 94.56&  93.70&80.94&  91.52\\
        \midrule
        \textbf{DPL (Ours)} &  \textbf{98.14} & \textbf{95.16} &  \underline{96.63} & \textbf{87.70} &  \textbf{94.41} \\
        \bottomrule
    \end{tabular}
    \vspace{-0.3cm}
\end{table}

\begin{table}[!tb]
    \centering
    \caption{\textbf{Cross-quality evaluation (AUC \%)}. Train set: FF++(\textbf{c23}). Test set: FF++(\textbf{c40}) with random JPEG compression.}
    \label{tab:cross_quality_evaluation}
    \vspace{-0.3cm}
    \setlength{\tabcolsep}{0.3cm}
    \begin{tabular}{lccccc}
        \toprule
        Methods & DF & F2F & FS & NT & Avg. \\
        \midrule
        \multicolumn{6}{c}{\textbf{Random JPEG compression on c40 test set}} \\
        \midrule
        F3Net (ECCV'20) \cite{qian2020thinking} & 90.41& 85.26& 89.02& \underline{73.80} &84.62\\
        SRM (CVPR'21) \cite{srm} & 90.87& 85.97& 90.89&68.24 &83.99\\
        SPSL (CVPR'21) \cite{spsl} & 89.03& 86.82& 88.50&70.03 &83.60\\
        SIA (ECCV'22) \cite{sia_eccv2022} & 89.05& \underline{86.96} & 89.57&  68.16&83.43\\
        RECCE (CVPR'22) \cite{recce_cvpr2022} & 90.09& 85.45& 90.51&68.97 &83.75\\
        UIA-ViT (ECCV'22) \cite{uia-vit_eccv2022} & 92.85& 85.73& 90.40&  \textbf{74.03} & \underline{85.75} \\
        QAD (ICCV'23) \cite{QAD_iccv2023} & 90.46& 84.30& \underline{91.83} &64.21 &82.70\\
        UCF (ICCV'23) \cite{ucf_iccv2023} & \underline{92.92} & \textbf{87.28} & 89.33&68.61 &84.53\\
        \midrule
        \textbf{DPL (Ours)} & \textbf{94.22} & 86.26& \textbf{92.88} &72.09& \textbf{86.36} \\
        \bottomrule
    \end{tabular}
    \vspace{-0.5cm}
\end{table}

\begin{table}[!tb]
    \caption{\textbf{Cross-quality with cross-manipulation evaluation (AUC \%).} Train set: FF++(c23). Test set: both FF++(c23) and FF++(c40) with random JPEG compression.}
    \label{tab:cross_manipulation_evaluation}
    \vspace{-0.3cm}
    \centering
    \begin{tabular}{c@{\hspace{0.4cm}}c@{\hspace{0.4cm}}c@{\hspace{0.4cm}}c@{\hspace{0.4cm}}c@{\hspace{0.4cm}}c@{\hspace{0.4cm}}c}
        
        \toprule
        Methods & Train & NT& F2F& FS & DF& Avg.\\
        \midrule
        \multicolumn{7}{c}{\textbf{Random JPEG Compression on c23 and c40 test set}} \\
        \midrule
        \multirow{4}{*}{QAD (ICCV'23) \cite{QAD_iccv2023}} & NT & 84.10& 63.70& 45.73& 72.34& \multirow{4}{*}{\underline{68.11}} \\
        & F2F & 55.31& 96.07& 56.28& 66.62& \\
        & FS & 48.66& 58.49& 98.34& 62.45& \\
        & DF & 62.18& 63.74& 56.39& 99.50& \\
        \midrule
        \multirow{4}{*}{UCF (ICCV'23) \cite{ucf_iccv2023}} & NT & 81.54& 60.24& 46.42& 64.14& \multirow{4}{*}{66.56}\\
        & F2F & 53.80& 94.87& 57.37& 64.78& \\
        & FS & 49.06& 55.15& 97.37& 67.77& \\
        & DF & 58.15& 58.80& 56.69& 99.03& \\
        \midrule
        \multirow{4}{*}{\textbf{DPL (Ours)}} & NT & 84.72& 62.41& 47.72& 70.14& \multirow{4}{*}{\textbf{68.97}}\\
        & F2F & 54.32& 95.33& 60.95& 69.04& \\
        & FS & 47.15& 58.40& 98.10& 74.20& \\
        & DF & 60.47& 63.16& 58.27& 99.18& \\
        \bottomrule
        
    \end{tabular}
\end{table}

\begin{table}[!tb]
    \caption{\textbf{Cross-quality with cross-dataset evaluation (AUC \%)}. Train set: FF++(c23). Test set: individual test with random JPEG compression.}
    \vspace{-0.3cm}
    \label{tab:cross_dataset_evaluation}
    \centering
    \setlength{\tabcolsep}{0.2cm}
    \begin{tabular}{lcccc}
        \toprule
        Methods & FSH & CDFv2&FFIW10k &Avg\\
        \midrule
        \multicolumn{5}{c}{\textbf{Random JPEG Compression on individual test set}} \\
        \midrule
        F3Net (ECCV'20) \cite{qian2020thinking} & 70.37& 67.41&65.85&67.78\\
        SRM (CVPR'21) \cite{srm} & 66.76& 66.40&66.81&65.98\\
        SPSL (CVPR'21) \cite{spsl} & 65.80& 69.34&65.74&65.74\\
        SIA (ECCV'22) \cite{sia_eccv2022} & 64.37& 64.78&  63.68&63.90\\
        RECCE (CVPR'22) \cite{recce_cvpr2022} & 66.92& 68.25&64.64&65.72\\
        UIA-ViT (ECCV'22) \cite{uia-vit_eccv2022} & \underline{72.60}& 70.52&  67.55& \underline{70.11} \\
        QAD (ICCV'23) \cite{QAD_iccv2023} & 67.51& \underline{70.58} & \underline{68.16} & 67.30\\
        UCF (ICCV'23) \cite{ucf_iccv2023} & 66.75& 66.89&64.92&65.64\\
        \midrule
        \textbf{DPL (Ours)} & \textbf{74.91} & \textbf{71.00} & \textbf{68.77} & \textbf{70.69} \\
        \bottomrule
    \end{tabular}
\end{table}

\smallskip
\noindent\textbf{Cross-quality with Cross-manipulation Evaluation.}
We also perform a coss-quality cross-manipulation evaluation on the FF++ dataset. Each method is trained on one manipulation type and tested across all four types (NT, F2F, FS, and DF). Given that QAD and UCF are among the most recent and effective methods, we compare them with our method. As shown in \cref{tab:cross_manipulation_evaluation}, our method still outperforms others on average, demonstrating its effectiveness in detecting varying qualities under different scenarios.

\smallskip
\noindent\textbf{Cross-quality with Cross-dataset Evaluation.} Furthermore, we perform a cross-quality cross-dataset evaluation, where all methods are trained on FF++(c23) and directly evaluated on the test set of the unseen dataset with random JPEG compression.
\cref{tab:cross_dataset_evaluation} shows the results under this setting. It can be seen that our method achieves the best performance on all datasets and averagely improves the AUC score by $0.58 \%$ compared to the second-best method UIA-ViT. This experiment further corroborates the practical feasibility of our method even across various datasets. 

\subsection{Ablation Study}
This section studies the effect of different settings in our method. Note that we apply random JPEG compression on the test set in each setting as in the main experiments.

\smallskip \noindent\textbf{Effect of Each Component.}
This part studies the effect of each component. Specifically, we study five settings: 1) \textbf{Baseline}: only employing the backbone network without DPL. 2) \textbf{w/o VQPL}: disabling the VQPL branch. 3) \textbf{w/o FIPL}: disabling the FIPL branch. 4) \textbf{w/o Train Stage II}: only applying random sampling $l_t$ in FSM. 5) \textbf{w/o $\mathcal{L}_{\textrm{REG}}$}: only using the cross-entropy loss $\mathcal{L}_{\textrm{CE}}$ without regularization loss $\mathcal{L}_{\textrm{REG}}$ in Stage I.
As shown in \cref{tab:different_component_evaluation}, without either VQPL or FIPL, the performance drops at FF++(c40) and FF++(c23), demonstrating their role in dual progressive learning branches. Moreover, omitting train stage II also drops the performance, corroborating the efficacy of network $\Phi$ in the feature selection module. Additionally, without $\mathcal{L}_{\textrm{REG}}$ introduces a certain performance drop, showing the effectiveness of this objective.

\smallskip \noindent\textbf{Effect of Different Backbones.}
\cref{tab:different_backbone_evaluation} shows the comparability of DPL with four different backbone networks: ConvNeXt-B \cite{convnext}, ConvNeXt-S \cite{convnext}, Swin-T \cite{liu2021Swin}, ResNet-50 \cite{resnet_cvpr2016}. The results show that the DPL improves the performance from 0.41$\%$ to 1.18$\%$ on average on two sets compared to the baselines.

\begin{table}[!t]
    \begin{minipage}[c]{0.5\textwidth}
    \centering
        \caption{Effect of each component.}
        \label{tab:different_component_evaluation}
        \vspace{-0.3cm}
        \centering
        \setlength{\tabcolsep}{0.2cm}
        \resizebox{\linewidth}{!}{\begin{tabular}{lcc}
            \toprule
            Setting & FF++(c40) & FF++(c23) \\
            \midrule
            Baseline & 84.65 &92.82  \\
            w/o VQPI&  85.68 &92.83  \\
            w/o FIPL&  86.31 &93.42  \\
            w/o Train Stage II &  86.32&94.39  \\
            w/o $\mathcal{L}_{\textrm{REG}}$ &  85.26&93.32  \\
            \midrule
            \textbf{DPL} & \textbf{86.36}&\textbf{94.41}  \\
            \bottomrule
        \end{tabular}}
    \end{minipage}
    \hfill
     \begin{minipage}[c]{0.47\textwidth}
             \centering
        \caption{Effect of different backbones.}
        \label{tab:different_backbone_evaluation}
        \vspace{-0.3cm}
        \setlength{\tabcolsep}{0.2cm}
        \resizebox{\linewidth}{!}{\begin{tabular}{lcc}
            \toprule
            Backbone & FF++(c40) & FF++(c23) \\
            \midrule
            ConvNeXt-B       & 86.43 & 93.61 \\
            ConvNeXt-B + DPL & 86.78 \blue{(+0.35)} & 94.61 \blue{(+1.00)} \\
            \midrule
            ConvNeXt-S & 85.05 & 93.35 \\
            ConvNeXt-S + DPL & 86.56 \blue{(+1.51)} & 94.20 \blue{(+0.85)} \\
            \midrule
            Swin-T & 85.98 & 93.20 \\
            Swin-T + DPL & 86.12 \blue{(+0.14)} & 93.44 \blue{(+0.24)} \\
            \midrule
            Res-50 & 84.38& 93.01 \\
            Res-50 + DPL & 84.86 \blue{(+0.48)} & 93.36 \blue{(+0.35)} \\
            \bottomrule
        \end{tabular}}
    \end{minipage}
\end{table}


\smallskip
\noindent \textbf{Various Fusion Strategies for Dual Branches.} This part studies three fusion strategies in \cref{tab:different_fusion_trategies}: 1) \textbf{Add}: directly adding the features from the VQPL branch and FIPL branch. 2) \textbf{Concat}: concatenating the features from the VQPL branch and FIPL branch. 3) \textbf{Attention}: merge the features from both branches using an attention module AFF \cite{AFF_wacv2021}. It can be seen that the strategy of directly adding features performs best from the overall effect.

\begin{table}[!t]
    \centering
    \begin{minipage}[c]{0.45\textwidth}
        \caption{Effect of different fusion strategies.}
        \vspace{-0.3cm}
        \setlength{\tabcolsep}{0.2cm}
        \centering
        \label{tab:different_fusion_trategies}
        \begin{tabular}{lcc}
            \toprule
            Setting & FF++(c40) & FF++(c23) \\
            \midrule
            Add  & 86.36& 94.41 \\
            Concat & 85.43& 94.26 \\
            Attention \cite{AFF_wacv2021} & 86.54& 93.46 \\
            \bottomrule
        \end{tabular}
    \end{minipage}
    \hfill
    \begin{minipage}[c]{0.45\textwidth}
        \caption{Effect of different settings of applying \(\mathcal{L}_{\textrm{REG}}\)}
        \vspace{-0.3cm}
        \setlength{\tabcolsep}{0.2cm}
        \centering
        \label{tab:different_apply_position}
        \begin{tabular}{lcc}
        \toprule
        Setting & FF++(c40) & FF++(c23) \\
        \midrule
        $\mathcal{L}_{\textrm{REG}}^*$ & 85.16& 92.41\\
        $\mathcal{L}_{\textrm{REG}}^{\dagger}$ & 86.20& 94.13\\
        $\mathcal{L}_{\textrm{REG}}^{\ddagger}$ & 86.35& 94.41\\
        \bottomrule
        \end{tabular}
    \end{minipage}
    \vspace{-0.3cm}
\end{table}


\smallskip
\noindent \textbf{Different Settings of Applying \(\mathcal{L}_{\textrm{REG}}\).} For the regularization loss  $\mathcal{L}_{\textrm{REG}}$, applying it to the output features at different time steps results in varying effects. \cref{tab:different_apply_position} presents three settings: 1) applying $\mathcal{L}_{\textrm{REG}}$ only to the output features of the first time step, denoted as $\mathcal{L}_{\textrm{REG}}^*$. 2) applying $\mathcal{L}_{\textrm{REG}}$ only to the output features of the maximum effective time step, denoted as $\mathcal{L}_{\textrm{REG}}^{\dagger}$. 3) applying $\mathcal{L}_{\textrm{REG}}$ to the output features of the maximum effective time step including all preceding time steps, denoted as $\mathcal{L}_{\textrm{REG}}^{\ddagger}$. The results show that the third strategy achieves the best performance.  



\subsection{Further Analysis}

\begin{figure}[tb]
    \centering
    \includegraphics[width=1\linewidth]{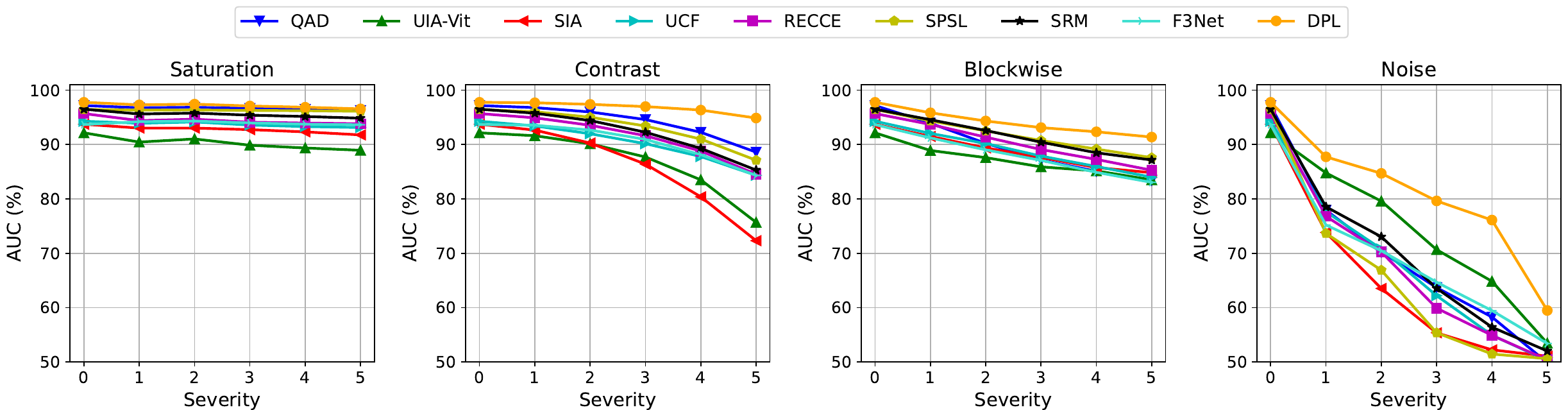}
    \caption{Performance of different methods under four perturbations.}
    \label{fig:roubustness}
    \vspace{-0.4cm}
\end{figure}

\smallskip
\noindent\textbf{Robustness Analysis.}
This part analyzes the robustness of our method to unseen perturbations. Specifically, we employ four perturbations: Saturation (color saturation change), Contrast (color contrast change), Blockwise (local block-wise distortion), Noise (white Gaussian noise in color components). Each perturbation has five severity levels.
\cref{fig:roubustness} shows the performance of different method against perturbations. It can be seen that our method consistently outperforms other methods across all scenarios.

\begin{wraptable}[7]{r}{0.5\textwidth}
    \caption{Performance of integrating CLIP image encoder for detection.}
    \setlength{\tabcolsep}{0.2cm}
    \centering
    \label{tab:integrate_clip_image_encoder}
    \resizebox{\linewidth}{!}{\begin{tabular}{lcc}
        \toprule
        Setting & FF++(c40) & FF++(c23) \\
        \midrule
        Concat(CLIP, $\Psi$) & 85.68& 92.96 \\
        Add(CLIP, $\Psi$) & 85.11& 94.01 \\
        Add(CLIP, $\mathcal{P}_1,\mathcal{P}_2$) & 85.12& 94.36 \\
        \bottomrule
    \end{tabular}
    }
\end{wraptable}

\smallskip
\noindent\textbf{Integrating CLIP for Detection.} Since CLIP contains powerful semantic priors, we explore integrating it for feature enhancement. Specifically, we investigate three settings: 1) Concatenating feature from CLIP image encoder with backbone $\Psi$. 2) Adding feature from CLIP image encoder with backbone $\Psi$. 3) Adding feature from CLIP image encoder with dual branches $\mathcal{P}_1,\mathcal{P}_2$. However, as shown in \cref{tab:integrate_clip_image_encoder}, integrating CLIP into DPL achieves inconspicuous improvement.

\begin{wrapfigure}[8]{r}{0.4\textwidth}
\vspace{-1cm}
    \centering
    \includegraphics[width=1\linewidth]{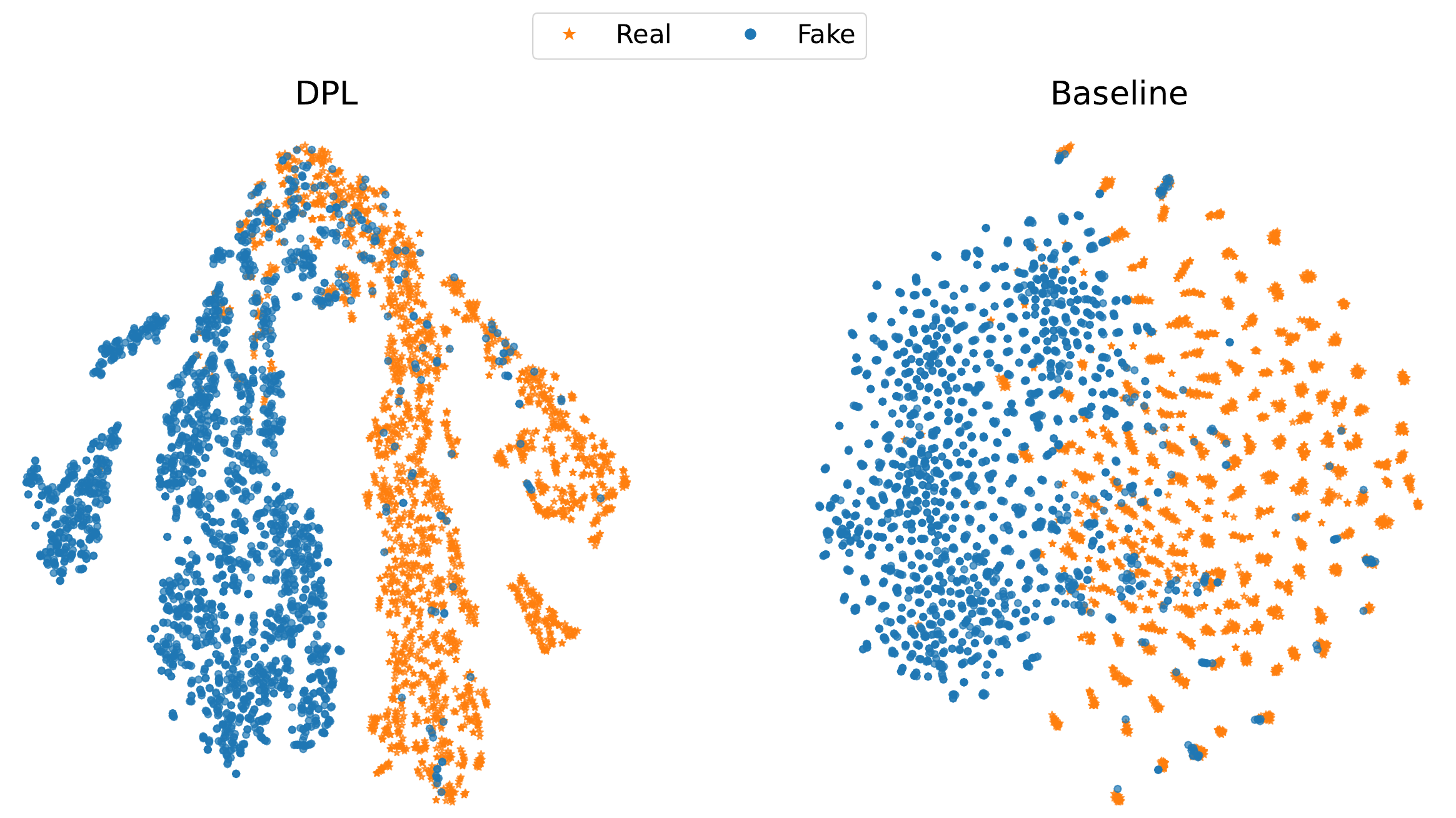}
    \caption{t-SNE visualization of DPL and baseline.}
    \label{fig:tsne}
\end{wrapfigure}

\begin{figure}[!t]
    \centering
    \includegraphics[width=0.95\linewidth]{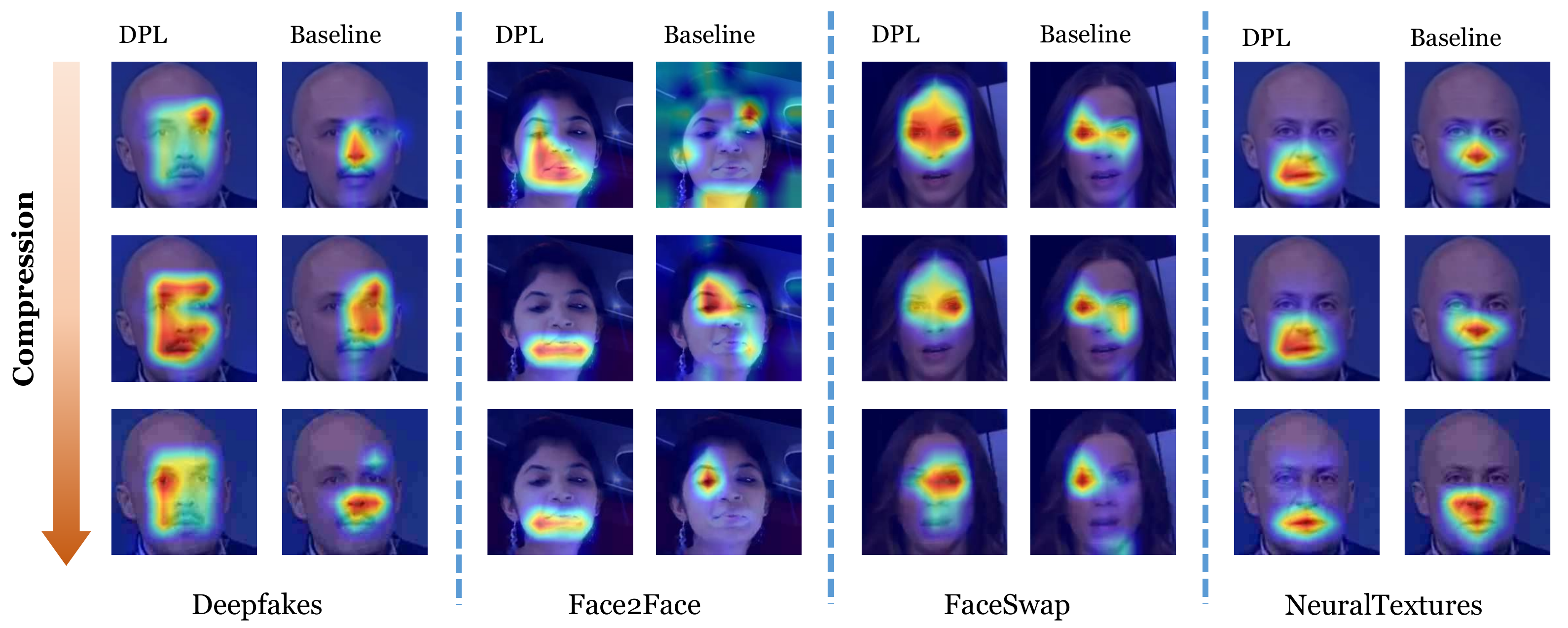}
    \caption{Grad-CAM visualization of DPL and baseline.}
    \label{fig:cam}
    \vspace{-0.4cm}
\end{figure}

\smallskip
\noindent\textbf{Attention Visualization.} We further use GradCAM \cite{grad_cam} to localize the activated regions to detect forgery. The visualization results on Deepfakes, Face2Face, FaceSwap, and NeuralTextures are shown in \cref{fig:cam}. It can be observed that, compared to the baseline, our method shows relatively stable attention regions on the forgery artifacts, demonstrating superiority against the compression.

\begin{figure}[!t]
    \begin{minipage}[c]{0.5\textwidth}
        \centering
        \includegraphics[width=0.9\linewidth]{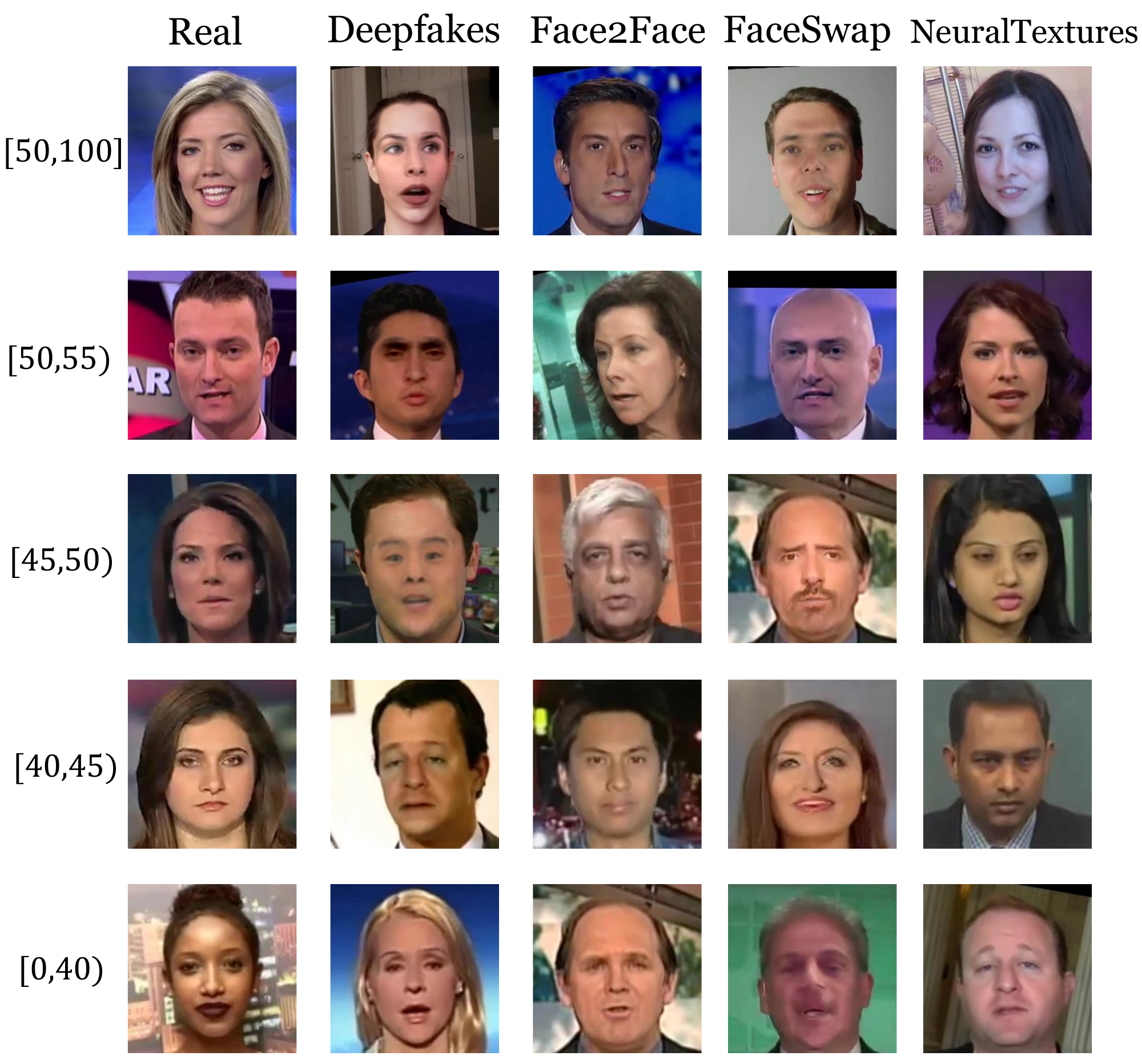}
        \caption{Analysis Result of VQI on FF++ (c23) dataset}
        \label{fig:VQI_example}
    \end{minipage}
    \hfill
    \begin{minipage}[c]{0.45\textwidth}
        \centering
        \includegraphics[width=0.9\linewidth]{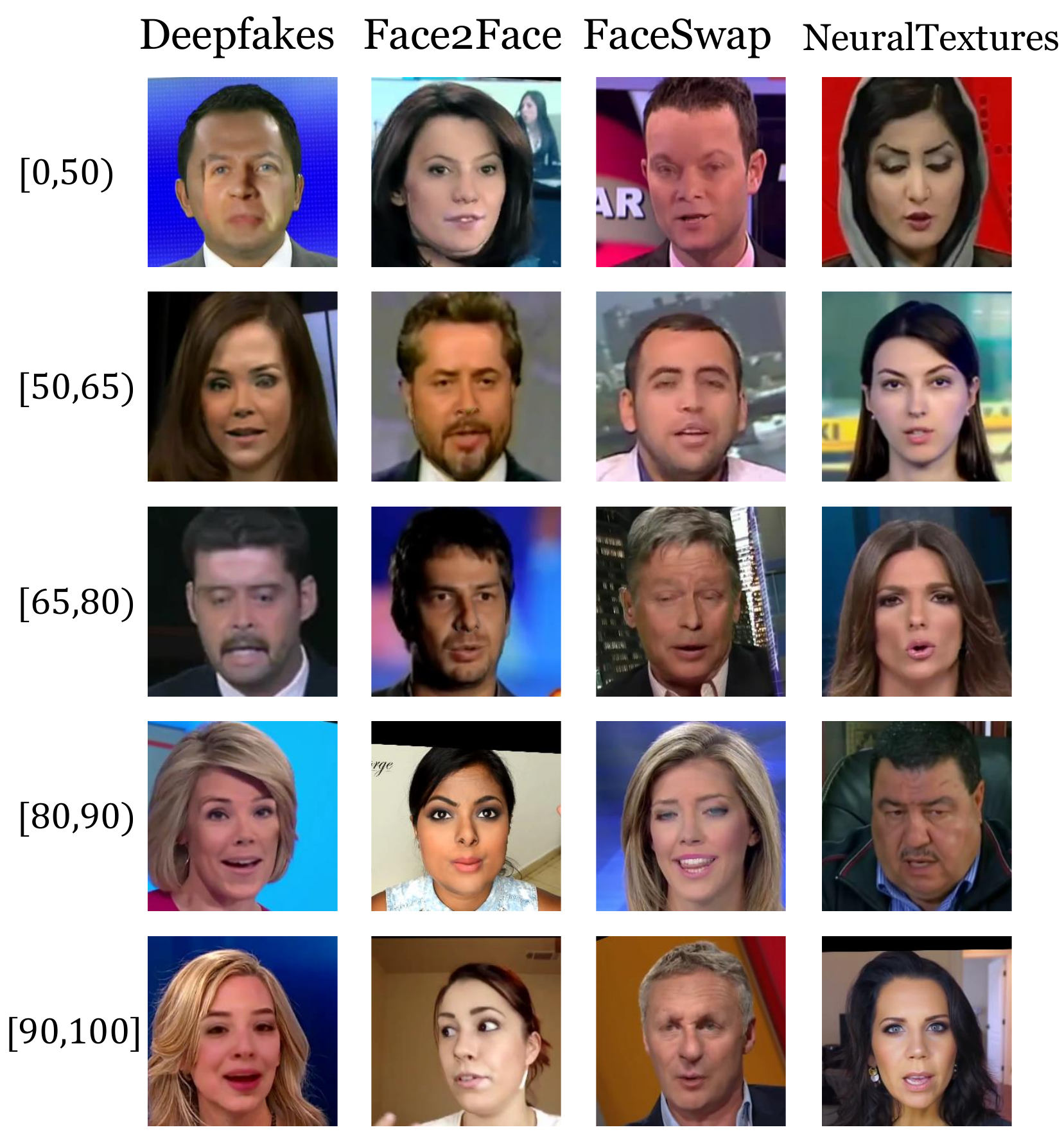}
        \caption{Analysis Result of FII on FF++ (c23) dataset}
        \label{fig:FII_example}
    \end{minipage}
    \vspace{-0.6cm}
\end{figure}

\smallskip
\noindent\textbf{Feature Visualization.} We visualize the fused feature from dual branches using t-SNE algorithm~\cite{tsne}. \cref{fig:tsne} shows the visual comparison between DPL and baseline method on FF++ (c23), demonstrating that our method effectively partition these samples into two different clusters while retaining less overlap.

\smallskip
\noindent\textbf{More Analysis on VQI.} We show several examples of different ranges of video quality in \cref{fig:VQI_example}. Note that we directly employ the prompt text in \cite{clip_iqa_iccv2023}. Specifically, we divide VQI score into different intervals: [50,100], [50,55), [45,50), [40,45), [0,40), with lower scores indicating worse image quality. These intervals correspond to the $K_1$ quality labels described in \cref{sec:vqi}. It can be observed that VQI can effectively distinguish images of different quality.


\smallskip
\noindent\textbf{More Analysis on FII.} We also show several examples of forgery identifiability in \cref{fig:FII_example}. The FII score is divided into five intervals as well, which are [0,50), [50,65), [65,80), [80,90), [90,100]. Moreover, we have attempted several paired text prompts, and inspected if the indication score matches the degree of manipulation.

\vspace{-0.3cm}
\section{Conclusion}
\vspace{-0.2cm}
This paper introduces a new Dual Progressive Learning (DPL) framework for cross-quality DeepFake detection. In contrast to existing methods, we analog this task to the progressive effort of ``drilling for underground water''. Our method involves two sequential-based branches designed to excavate the forgery traces based on the levels of video quality and forgery identifiability, estimated by dedicated CLIP-based indicators. These branches share a feature selection module that prepares tailored features for different time steps. To train this framework, we propose a two-stage pipeline using the PPO optimization with dedicated objectives. Experimental results validate the effectiveness of our method in detecting DeepFakes across varying video qualities.

\smallskip
\noindent\textbf{Acknowledgement: }
This work is supported in part by the National Natural Science Foundation of China (No.62402464), Shandong Natural Science Foundation (No.ZR2024QF035), China Postdoctoral Science Foundation (No.2021TQ0314, No.2021M703036).
 

%
%
\bibliographystyle{splncs04}
\bibliography{main}
\end{document}